\documentclass{article}

\usepackage{arxiv}

\usepackage{multirow}
\usepackage{array}
\usepackage[ruled,vlined]{algorithm2e}

\usepackage{xcolor}

\usepackage[utf8]{inputenc} 
\usepackage[T1]{fontenc}    
\usepackage{hyperref} 
\usepackage{scalerel,stackengine}
\stackMath
\newcommand\reallywidehat[1]{%
\savestack{\tmpbox}{\stretchto{%
  \scaleto{%
    \scalerel*[\widthof{\ensuremath{#1}}]{\kern-.6pt\bigwedge\kern-.6pt}%
    {\rule[-\textheight/2]{1ex}{\textheight}}
  }{\textheight}%
}{0.5ex}}%
\stackon[1pt]{#1}{\tmpbox}%
}

\usepackage{graphicx}
\usepackage[export]{adjustbox}
\usepackage{xcolor}
\usepackage{subfigure}
\usepackage{colortbl}
\usepackage{paralist}
\usepackage{bbm}
\usepackage{url}            
\usepackage{booktabs}       
\usepackage{amsfonts}       
\usepackage{nicefrac}       
\usepackage{microtype}      
\usepackage{lipsum}	
\usepackage{xcolor}
\title{Are Chess Discussions Racist? An Adversarial Hate Speech Data Set}

\author{
  Rupak Sarkar\\
  \small{Maulana Abul Kalam Azad University of Technology}\\
  \texttt{rupaksarkar.cs@gmail.com } \\
\And
Ashiqur R. KhudaBukhsh \\
  \small{Carnegie Mellon University}\\
  \texttt{akhudabu@cmu.edu} \\
}

\begin{document}
\maketitle

\begin{abstract}

On June 28, 2020, while presenting a chess podcast on Grandmaster Hikaru Nakamura, Antonio Radi\'c's YouTube handle got blocked because it contained ``harmful and dangerous'' content. YouTube did not give further specific reason, and the channel got reinstated within 24 hours. However, Radi\'c speculated that given the current political situation, a referral to ``black against white'', albeit in the context of chess, earned him this temporary ban. In this paper, via a substantial corpus of 681,995 comments, on 8,818 YouTube videos hosted by five highly popular chess-focused YouTube channels, we ask the following research question: \emph{how robust are off-the-shelf hate-speech classifiers to out-of-domain adversarial examples?} We release a data set of 1,000 annotated comments\footnote{This paper is accepted at AAAI 2021 (student abstract). The data set is available at \url{https://www.cs.cmu.edu/~akhudabu/Chess.html}.}  where existing hate speech classifiers misclassified benign chess discussions as hate speech. We conclude with an intriguing analogy result on racial bias with our findings pointing out to the broader challenge of color polysemy.

\end{abstract}


\section{Introduction}
On June 28, 2020, while presenting a chess podcast on Grandmaster Hikaru Nakamura, Antonio Radi\'c's YouTube handle (Agadmator's Chess Channel) got blocked because it contained ``harmful and dangerous'' content. The channel got reinstated in 24 hours, and YouTube didn't provide any specific reason for this temporary ban. However, Radi\'c speculated that under the current political circumstances\footnote{\url{https://www.bbc.com/news/world-us-canada-53273381}}, a referral to ``black against white'' in a completely different context of chess, cost him the ban\footnote{\url{https://www.thesun.co.uk/news/12007002/chess-champ-youtube-podcast-race-ban/}}. The swift course-correction by YouTube notwithstanding, in the age of AI monitoring and filtering speech over the internet, this incident raises an important question: \emph{is it possible that current hate speech classifiers may trip over benign chess discussions, misclassifying them as hate speech?} If yes, how often does that happen and is there any general pattern to it? In this paper, via a substantial corpus of 681,995 comments on 8,818 YouTube videos hosted by five highly popular chess-focused YouTube channels, we report our ongoing research on adversarial examples for hate speech detectors.

Hate speech detection, a widely-studied research challenge, seeks to detect communication disparaging a person or a group on the basis of race, color, ethnicity, gender, sexual orientation, nationality, religion, or other
characteristics~\cite{nockleby2008hate}. Hate speech detection research in various social media platforms such as Facebook~\cite{del2017hate}, Twitter~\cite{davidson2017automated} and YouTube~\cite{dinakar2012common} has received a sustained focus. While the field has made undeniable progress, the domain-sensitivity of hate speech classifiers~\cite{arango2019hate} and susceptibility to adversarial attacks~\cite{grondahl2018all} are well-documented. 

In this paper, we explore the domain of online chess discussions where \colorbox{blue!30}{white} and \colorbox{blue!30}{black} are always adversaries; \colorbox{blue!30}{kills}, \colorbox{blue!30}{captures}, \colorbox{blue!30}{threatens}, and \colorbox{blue!30}{attacks} each other's pieces; and terms such as  \colorbox{blue!30}{Indian} defence, Marshall \colorbox{blue!30}{attack} are common occurrences. Our primary contribution is an annotated data set of 1,000 comments verified as \textbf{not} hate speech by human annotators that are incorrectly flagged as hate speech by existing   classifiers.

\section{Data Set and Hate Speech Classifiers}

We consider five chess-focused  YouTube channels listed in Table~\ref{tab:channels}. We consider all videos in these channels uploaded on or before July 5, 2020, and use the publicly available YouTube API to obtain comments from these 8,818 videos. Our data set consists of 681,995 comments (denoted by $\mathcal{D}_\mathit{chess}$) posted by 172,034 unique users. 

\begin{table}[htb]
{
\scriptsize
\begin{center}
     
\begin{tabular}{|l | c | c | c |}
    \hline
    YouTube handles & \#Subscribers & \#Videos & \#Comments $ $\\
    \hline                                 
   Agadmator's Chess Channel  &  772k & 1,780 & 420,350 \\ 
    \hline
   MatoJelic & 143k & 2,976 & 126,032  \\
    \hline
   Chess.com  & 388k & 2,189 & 61,472 \\
    \hline
   John Bartholomew  & 140k & 1,619 & 589,38 \\
    \hline
   GM Benjamin Finegold  &  59.3k & 254 & 15,203 \\
    \hline
    \end{tabular}

\vspace{0.5cm}
\caption{List of YouTube channels considered.}
\label{tab:channels}
\end{center}
}
\end{table}

\begin{table}[htb]
{
\scriptsize
\begin{center}
     
\begin{tabular}{|l| c | c | c |}
    \hline
     & $\mathcal{M}_\mathit{twitter}$ 
       & $\mathcal{M}_\mathit{stormfront}$ \\
     &\cite{davidson2017automated} &\texttt{BERT} trained on~\cite{gibert2018hate}\\
    \hline                                 
   Fraction of positives   &  1.25\% &  0.43\% \\ 
    \hline
    Human evaluation  &  5\% (true positive) &  15\% (true positive) \\
  on predicted positives & 87\% (false positive)  & 82\% (false positive)  \\
     & 8\% (indeterminate) & 3\% (indeterminate)   \\
    \hline
    \end{tabular}

\vspace{0.5cm}
\caption{Classifier performance on $\mathcal{D}_\mathit{chess}$. }
\label{tab:results}
\end{center}
}
\end{table}

\begin{table}[htb]
{
\scriptsize
\begin{center}
     \begin{tabular}{|p{0.8\textwidth}|}
     \hline
\textbf{\textcolor{black}{\texttt{That is one of the most beautiful attacking sequences I have ever seen, black was always on the back foot. Thank you for sharing. Seeing your channel one day in my recommended got me playing chess again after 15 years. All the best.}}}\\     

\hline
\textbf{\texttt{At 7:15 of the video Agadmator shows what happens when Black goes for the queen. While this may be the most interesting move, the strongest continuation for Black is Kg4. as Agadmator states, White is still winning. But Black can prolong the agony for quite a while.}} \\
\hline 
\textbf{\texttt{White's attack on Black is brutal. White is stomping all over Black's defenses. The Black King is gonna fall\ldots}}\\
    \hline
\textbf{\texttt{That end games looks like a draw to me... its hard to see how it's winning for white. I seems like black should be able to block whites advance.}}  \\  
    \hline
\textbf{\texttt{\ldots he can still put up a fight (i dont see any immediate threat black can give white as long as white can hold on to the e-rook)
}} \\
    \hline 
    \end{tabular}
    
\end{center}
\vspace{0.5cm}
\caption{{Random samples of misclassified hate speech.}}
\label{tab:hateSpeech}}
\end{table}


We consider two hate speech classifiers: \textcolor{blue}{(1)} an off-the-shelf hate speech classifier~\cite{davidson2017automated} trained on twitter data (denoted by $\mathcal{M}_\mathit{twitter}$); and a \textcolor{blue}{(2)} a \texttt{BERT}-based classifier trained on annotated data from a white supremacist forum~\cite{gibert2018hate} (denoted as $\mathcal{M}_\mathit{stormfront}$). 

\section{Results}

We run both classifiers on $\mathcal{D}_\mathit{chess}$. We observe that only a minuscule fraction of comments are flagged as hate speech by our classifiers (see, Table~\ref{tab:results}). We next manually annotate randomly sampled 1,000 such comments marked as hate speech by at least one or more classifiers. Overall, we obtain 82.4\% false positives, 11.9\% true positives, and 5.7\% as indeterminate. High false positive rate indicates that without a human in the loop, relying on off-the-shelf classifiers' predictions on chess discussions can be misleading. We next evaluate individual false positive performance by manually annotating 100 randomly sampled  comments marked as hate speech by each of the classifiers. We find that~$\mathcal{M}_\mathit{twitter}$ has slightly higher false positive rate than $\mathcal{M}_\mathit{stormfront}$. Also, $\mathcal{M}_\mathit{stormfront}$ flags substantially fewer comments as hate speech than $\mathcal{M}_\mathit{twitter}$. Since $\mathcal{M}_\mathit{stormfront}$ is trained on a white supremacist forum data set, perhaps this classifier has seen hate speech targeted at the black community on a wider range of contexts. Hence, corroborating to the well-documented domain-sensitivity of hate speech classifiers, $\mathcal{M}_\mathit{stormfront}$ performs slightly better than  $\mathcal{M}_\mathit{twitter}$ trained on a more general hate speech twitter data set. Table~\ref{tab:hateSpeech} lists a random sample of false positives from $\mathcal{M}_\mathit{stormfront}$ and $\mathcal{M}_\mathit{twitter}$. We note that presence of words such as \texttt{black}, \texttt{white}, \texttt{attack}, \texttt{threat} possibly triggers the classifiers.

\section{Black is to Slave as White is to?}

We conclude our paper with a simple yet powerful observation.  Word associations test (e.g.,  \emph{France}:\emph{Paris}::\emph{Italy}:\emph{Rome}) on Skip-gram embedding spaces~\cite{mikolov2013distributed} are well-studied. Social biases in word embedding spaces is a well-established phenomenon~\cite{garg2018word} with several recent lines of work channelised to debiasing efforts~\cite{manzini-etal-2019-black}. We perform the following analogy test: \emph{black}:\emph{slave}::\emph{white}:\emph{?}, on $\mathcal{D}_\mathit{chess}$ and two data sets containing user discussions on YouTube videos posted on official channels of Fox News ($\mathcal{D}_\mathit{fox}$)  and CNN ($\mathcal{D}_\mathit{cnn}$) in 2020. While both $\mathcal{D}_\mathit{fox}$ and $\mathcal{D}_\mathit{cnn}$ predict \emph{slavemaster}, $\mathcal{D}_\mathit{chess}$ predicts \emph{slave}! Hence, the \colorbox{blue!30}{captures}, \colorbox{blue!30}{fights}, \colorbox{blue!30}{tortures} and \colorbox{blue!30}{killings} notwithstanding, over the 64 black and white squares, the two colors attain a rare equality the rest of the world is yet to see.


\bibliographystyle{unsrt} 

\end{document}